\documentclass[letterpaper, 10 pt, conference]{ieeeconf}  

\IEEEoverridecommandlockouts                              
\usepackage[dvips]{graphicx}
\usepackage{verbatim}

\usepackage{xcolor}

\usepackage{amsmath, amssymb}
\usepackage{bm} 
\usepackage{hyperref}

\usepackage{dblfloatfix}    

\usepackage{graphics} 
\usepackage{epsfig} 

\definecolor{my_gray}{gray}{0.8}

\def\yen{{\setbox0=\hbox{Y}Y\kern-.97\wd0\vbox{\hrule height.1ex
			width.98\wd0\kern.33ex\hrule height.1ex width.98\wd0\kern.45ex}}}

\title{\LARGE \bf
Visual Task Progress Estimation with Appearance Invariant Embeddings for Robot Control and Planning 
}

\author{Guilherme Maeda$^{1*\dag}$, Joni V\"{a}\"{a}t\"{a}inen$^{2\dag}$, Hironori Yoshida$^{1}$%
\thanks{ $^1$Preferred Networks Inc. 1-6-1 Otemachi, Chiyoda, Tokyo, Japan. {\tt\footnotesize  gjmaeda@preferred.jp, hyoshida@preferred.jp} 
}
\thanks{$^2$Faculty of Science and Engineering, Dept. of Modern Mechanical Engineering, Waseda University, Tokyo, Japan. 
This work is an achievement during internship at Preferred Networks.
		{\tt\footnotesize joni.vaatainen@moegi.waseda.jp}
}
\thanks{$^*$Corresponding author. $\dag$ authors with equal contribution.
}
}

\begin{document}

\maketitle
 \thispagestyle{empty}
\pagestyle{empty}

\begin{abstract}
One of the challenges of full autonomy is to have a robot capable of manipulating its current environment to achieve another environment configuration. This paper is a step towards this challenge, focusing on the visual understanding of the task. Our approach trains a deep neural network to represent images as measurable features that are useful to estimate the progress (or phase) of a task. The training uses numerous variations of images of identical tasks when taken under the same phase index. The goal is to make the network sensitive to differences in task progress but insensitive to the appearance of the images. To this end, our method builds upon Time-Contrastive Networks (TCNs) to train a network using only discrete snapshots taken at different stages of a task. A robot can then solve long-horizon tasks by using the trained network to identify the progress of the current task and by iteratively calling a motion planner until the task is solved. We quantify the granularity achieved by the network in two simulated environments. In the first, to detect the number of objects in a scene and in the second to measure the volume of particles in a cup. Our experiments leverage this granularity to make a mobile robot move a desired number of objects into a storage area and to control the amount of pouring in a cup.    
\end{abstract}

\color{black}

\section{Introduction}

\begin{figure}
	\centering
	\vspace{0.283425cm}
	\includegraphics[width=0.98\linewidth]{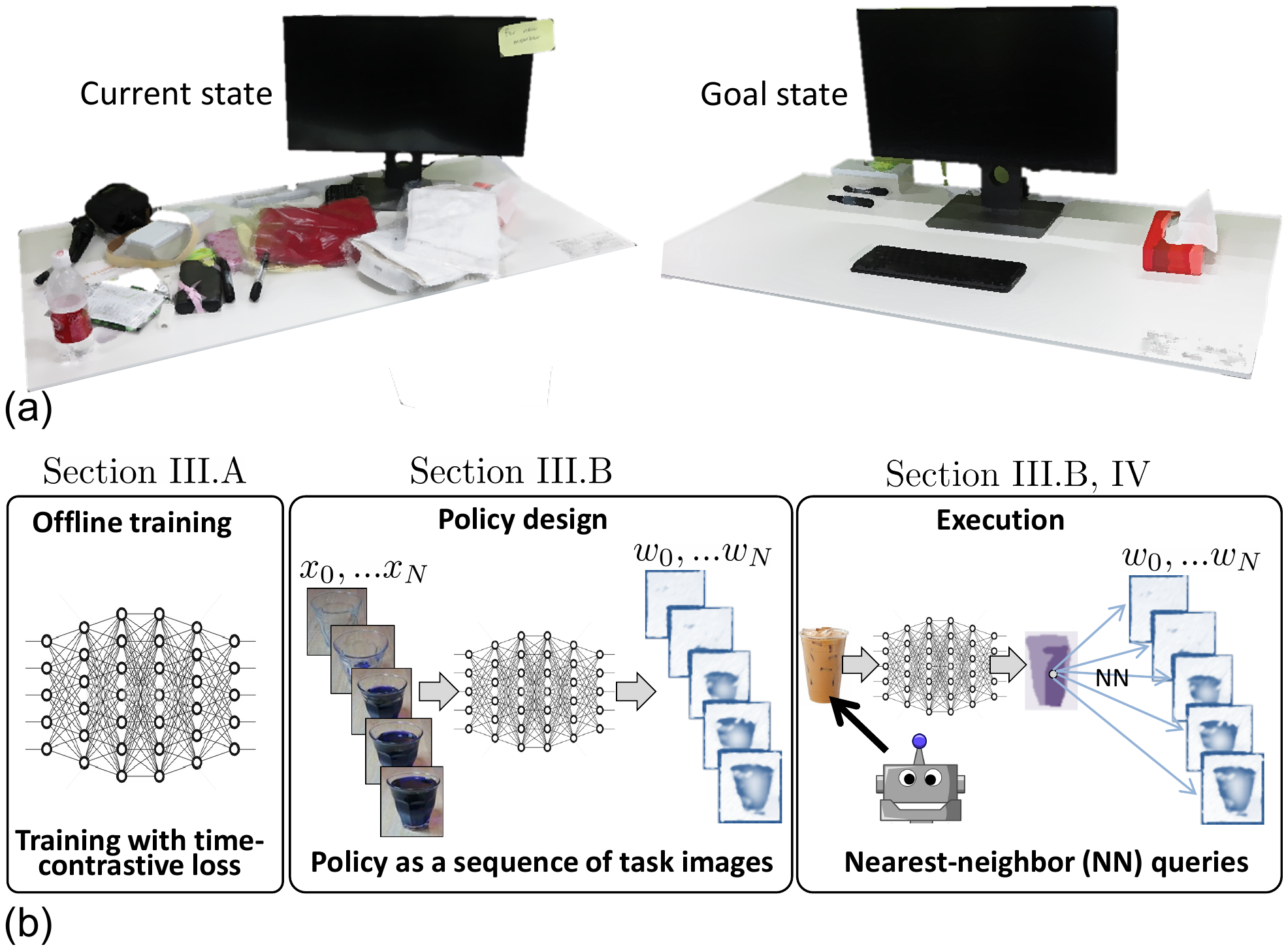}
	\caption{
		(a) A robot must be able to reason about the relative state of a disorganized desk concerning an organized desk. The result of this reasoning can be used in conjunction with a planner to produce long-horizons cleaning motor actions.
		To this end, we investigate a metric to estimate task progress that can be extracted directly from images.
		(b) The method comprises three steps. The offline training of the network. The generation of the policy from a sequence of images showing the gradual accomplishment of a task. And the execution of robot actions guided by the nearest-neighbor in the policy sequence.
	}  
	\vspace{-0.00cm}
	\label{fig:intro}
\end{figure}

One of the most basic capabilities of an intelligent autonomous system is to be able to reason about its current world, understand how it is different from a more desirable world, and act towards improvement by modifying its environment.
For example, in Figure \ref{fig:intro}(a), a service robot could use its camera to capture the image of a disorganized desk.
Using an appropriate metric, it would then compare this image with the one of a clean desk (say, retrieved from memory).
This difference is then transformed into a tidy up action, which may be realized by a series of manipulation commands.
This process is then repeated over a long horizon until the robot finds no discrepancy between the states of the images.

The main challenge in this paper is that of learning a network that is invariant to the appearance of images, but sensitive to the   progress.
The distance of the embedded images should be small for images that look dissimilar but represent the same phase (or progress) of the task.
For example, two different desks observed from different viewpoints that are both clean and organized are identical regarding the progress of a cleanup task.
Conversely, two images that look similar but represent different stages of the task should be identified as being far from each other.

As shown in Figure \ref{fig:intro}(b), our approach consists of three steps.
First, the offline training of a Time-Contrastive Network \cite{sermanet2018time} to learn the metric for progress estimation in a self-supervised fashion.
Once the network is trained, we generate a policy by feeding the network with a sequence of images $(x_0, ..., x_N)$ that shows the gradual accomplishment of a desired task, thus obtaining a sequence of embeddings $(w_0, ..., w_N)$. 
During execution, the robot can then use the same network to process its visual input and compare it with the sequence of embeddings obtained in the previous step. 
The nearest-neighbor in the sequence indicates the progress of the task and can be associated with a robot action.

Solving the invariance to appearance has many implications for the practicality of a service robot.
It allows the system to be trained on images from a multitude of environments regardless of the particular robot in use and the actual working environment (which is only precisely known after deployment),
it frees a mobile robot from the need to move to pre-defined positions from which to make observations,
it opens the opportunity to seamlessly combine real and simulated images as training data,
it supports lifelong learning of robots 
as scene and objects evolve/change but the underlying task goal remains; to cite a few.

The contribution of this paper is to introduce a method to allow a robot to estimate task progress  from vision by proposing a modified time-contrastive training based on TCNs \cite{sermanet2018time}. 
We also suggest a policy based on the network output to associate local actions to a sequence of images representing the gradual task accomplishment.
This policy allows a motion planner to solve long-horizon tasks (e.g. cleaning up a floor by moving blocks one-by-one), and can also be used as a feedback error signal.
With the support of simulated experiments, we discuss and provide an intuition of why the triplet-loss metric finds a suitable place in learning from tasks that have visually discernible phases and how this insight may be explored to increase efficiency in training.

\section{Related Work}

We discuss the background on the components and related research that form the basis for the design of our method.

\subsection{Goal as Images}

Extracting and using goals from images has been a long-standing challenge in robotics 
\cite{kuniyoshiLearningWatchingExtracting1994, bentivegnaLearningTasksObservation2004}
with past work reliant on the ingenuity of feature design for the processing of relevant information.
Recently, automatic feature and representation learning of goals from images with deep learning have become predominant.
Such goal representations have been not only used in imitation and reinforcement learning \cite{finnDeepSpatialAutoencoders2016, sermanetUnsupervisedPerceptualRewards2017, nairVisualReinforcementLearning2018} but also to transfer images of simulated goals to the real world with domain randomization \cite{Pinto-RSS-18}.

Procedurally close to our paper, but under a setting of grasping in clutter, Jang et al. \cite{jang18a_grasp2vec} proposed a method where the desired image of a grasped object is given as a goal, and the robot learns to compare dissimilarities of this image with the result of its actions.
However, as in most of the visuomotor learning research with real robots, such methods have been almost exclusively developed and validated on short-horizon tasks without a clear mechanism for long-term planning.
Moreover, the usual use of fixed arms robots (as opposed to mobile bases) means that the invariance to the viewpoint is usually not the main concern in such works.

\subsection{Uses of Contrastive Learning in Robotics}

Triplet loss, which is the core metric to train the network in this paper, is perhaps most known in face recognition \cite{schroffFacenetUnifiedEmbedding2015}.
This loss was motivated as a way to recognize faces by directly measuring distances in a continuous space of features without relying on explicit layers for classification. 
Deep convolutional networks trained under this loss have shown to achieved state-of-the-art performance \cite{hermansDefenseTripletLoss2017a}.

Contrastive losses have also found uses in the context of robotics, mainly for learning visual descriptors in both 2D (RGB) \cite{schmidtSelfsupervisedVisualDescriptor2016} and 3D (RGB-D) images \cite{zeng3dmatchLearningLocal2017}.
Such descriptors have many applications in robotics, from mapping and localization to navigation, and manipulation.
Dense descriptors have been proposed to track a specific keypoint across variations of the image \cite{schmidtSelfsupervisedVisualDescriptor2016} or poses of an object \cite{florenceDenseObjectNets2018}.

Although the use of contrastive learning in robotics has mainly focused on the perception problem per se, dense descriptors have a close link to robot control.
They can be used to provide a controller with a target for grasping \cite{florenceDenseObjectNets2018} or a set of compact descriptors \cite{florenceSelfSupervisedCorrespondenceVisuomotor2019} to learn a control policy.
Since these methods are designed for learning correspondences of points between images while we focus on measuring the distance to finish a task, the resulting methods are quite different but show the versatility of contrastive loss in learning features for robotics applications.

Time-Contrastive Networks (TCNs) \cite{sermanet2018time} introduced a self-supervised, time-based approach to the triplet loss training.
TCNs embeddings were used to learn motor skills from observations of humans \cite{sermanet2018time} in both imitation and reinforcement learning.
A multi-frame extension capable of capturing velocity concepts from images was used to learn polices for swing-up a cart-pole and for running a half-cheetah \cite{dwibedi2018learning} in simulation.
TCNs were also used as part of a sim-to-real adaptation technique where a robot first learned motor skills in simulation using deep reinforcement learning, and the time-based contrastive loss was added for transferring the task to the real robot \cite{jeongSelfSupervisedSimtoRealAdaptation2019}.
We discuss the relation of our work with TCNs in detail in Section \ref{sec:tcn2}.

\subsection{Grounding Actions via Motion Planning}

Our goal is to leverage motion planners---and task-and-motion planners (TAMP) for complex cases---to ground actions obtained from the network embeddings.
Since TCNs were originally proposed as a representation for visuomotor skill learning, one may wonder why not use the TCN to learn end-to-end.
Our first issue is that of practicality. Particularly due to the need for creating training data where the robot is part of the images.
This leads to networks that are not transferable to a different robot (unless training data with the new robot is fed into the system), the robot motion becomes dependent on how the task was demonstrated, and failed roll-outs must be demonstrated to provide useful signal during reinforcement learning.

For that matter, the work of Liu et al. \cite{liuImitationObservationLearning2018} where the robot learns the context of a task from multiple views, partially overcomes the issue of having the robot in the view by using tools with long handles (sticks, brushes, spatulas). 
As such, when observed from the limited field of view of the camera, it is not possible to discern who (robot or human) is manipulating the object.
Thus, while such an approach circumvents the issue of collecting training data in the presence of a robot, it also constrains the range of tasks and the means a robot is allowed to interact with the environment.

Certainly, while visuomotor learning with real robots is progressing at an unprecedented pace in both reinforcement learning (e.g.  \cite{levineEndtoendTrainingDeep2016, finnDeepSpatialAutoencoders2016}) and imitation learning (e.g. \cite{liuImitationObservationLearning2018, yuOneshotImitationObserving2018}), at the current stage, long horizon problems are still extremely hard to tackle with deep policies.
Besides, no single visuomotor policy addresses the full stack required by mobile manipulators\footnote{for example, SLAM and planning of the base navigation combined with task plans, object pose estimation, and grasping.}, and even when learning a single short-horizon grasping action, it is impractical to imagine a commercial service robot training its physical motor skills in a hospital or in a childcare facility. 

On the other hand, motion planners are widely applied in the real-world and commercial applications \cite{laumondKineoCAMSuccess2006, krogerOpeningDoorNew2011} and can leverage the ubiquity of RGB-D/LIDAR cameras and algorithms for instance segmentation and matching to obtain positional information.
Planners produce predictable, often verified solutions, and are supported by a wider, mature body of work.
For the sake of practicality and real-world deployment in the short term, until many of the issues with visuomotor control with deep policies are solved (in particular long-horizon TAMP problems \cite{dantamIncrementalTaskMotion2016}), we believe that recurring to existing planning methods is a reasonable trade-off, particularly when higher levels of technological readiness must be factored in.

\section{Learning an Appearance Invariant Metric of Task Accomplishment} \label{sec:tcn2}

We re-visit TCNs and propose to train the network to specialize in the degree of accomplishment of a goal rather than on robot skills.
In other words, we modify the TCN training to solve the question of \emph{what} {is the state of the task} rather than on \emph{how} {to achieve the task}.
The capability of TCNs to estimate task progress has not been investigated so far.
The closest evaluation in the spirit of our paper's goal was to classify the keyframes of a video during pouring in its different stages such as ``within pouring distance'', ``liquid is flowing'', ``recipient has liquid'', etc. \cite{sermanet2018time}.
Here, we push into this direction and investigate whether TCNs can achieve the understanding of the task progress at a fine-grained level, for example, to allow a robot to control the pouring of a liquid at any arbitrary level in a cup.

Recall from Figure \ref{fig:intro}(b) that the proposed pipeline consists of three steps.
This section will explain each of these steps in more detail.

\subsection{Time-Based Self-Supervision with Triplet Loss} \label{sec:methodTraining}

\begin{figure}
	\centering
	\vspace{0.283425cm}
	\includegraphics[width=0.9\linewidth]{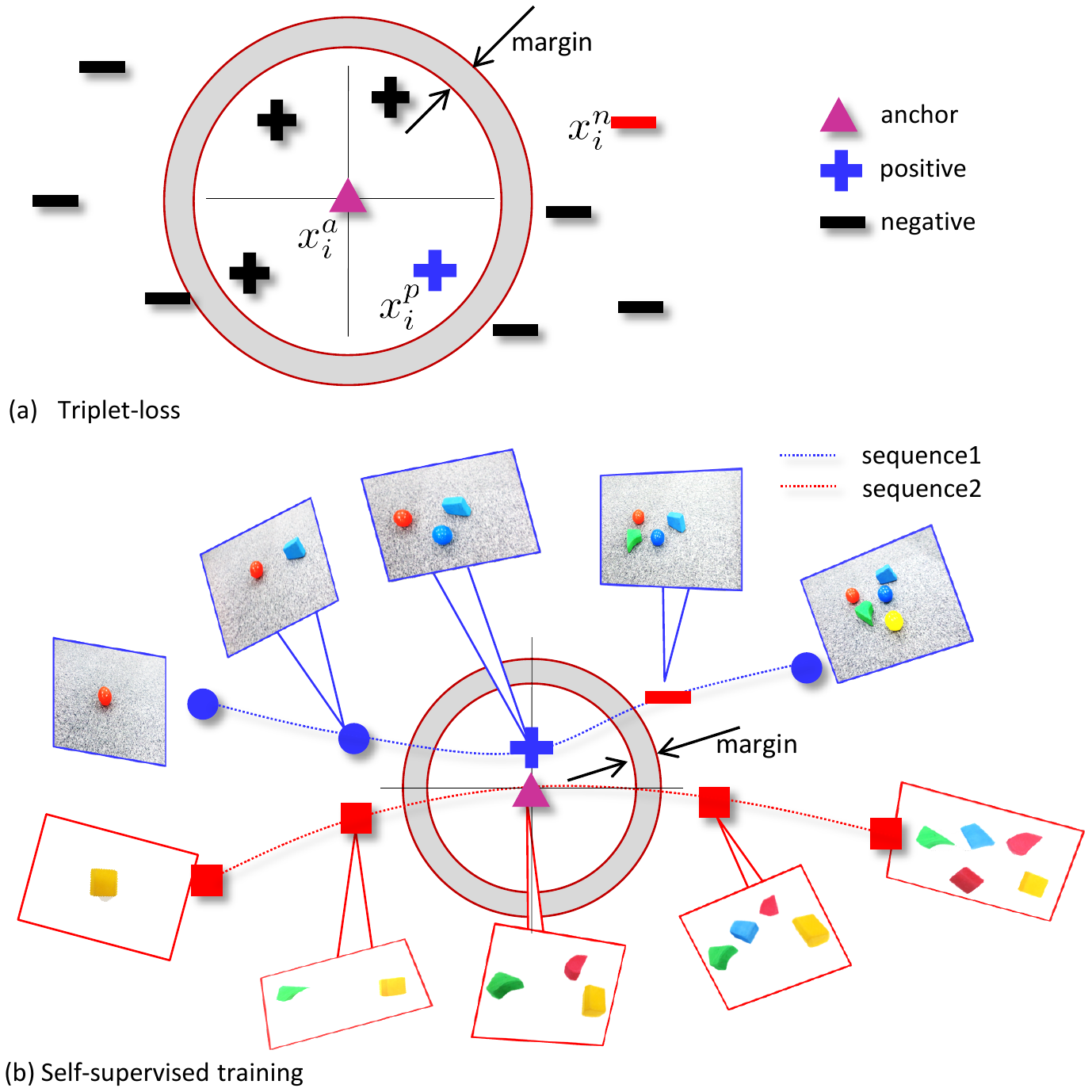}
	\caption{(a) A network trained on triplet loss separates positive and negative examples by learning to locate the margin over all possible triplets in the training data.
		(b) The separation of the triplet loss when using sequential data where each frame is taken at a different phase of the task. In this case, the adjacent frames are hard cases and one should expect the dissimilarity to increase as the phase difference relative to the anchor increases.}
	\label{fig:triplet_loss}
\end{figure}

Figure \ref{fig:triplet_loss}(a) illustrates the basic concept of a triplet loss as proposed by Schroff et al. \cite{schroffFacenetUnifiedEmbedding2015}.
The network is trained to encode high-dimensional images $x$  into a low-dimensional Euclidean space where the similarity between embeddings can be directly compared.
Thus, the network implements a function $f(x) \in  \mathbb{R}^d$, where the output is a d-dimensional vector.
Given a data set with $T$ possible triplets, for a given instance $i \in \{1,...,T\}$ an image $x_i^a$ is selected as the anchor, an image belonging to the same class of the anchor is selected as a positive $x_i^p$, and an image of a different class is selected as a negative $x_i^n$.
The network is trained by sampling from the set of all possible image triplets to minimize the cost
\begin{equation} \label{eq:tripletloss}
\sum_i^T   \left[ ||f(x_i^a)-f(x_i^p)||^2_2  - ||f(x_i^a)-f(x_i^n)||^2_2 + \alpha \right],
\end{equation}
where $\alpha$ is a margin of separation.

Figure \ref{fig:triplet_loss}(b) illustrates the use of triplet loss to quantify distances between phases of task sequences following the time contrastive training formulation proposed in \cite{sermanet2018time}.
Assume two sequences of images where the images are phase-aligned\footnote{we emphasize the use of phase rather than time as time-alignment does not imply that the different signs of progress among tasks are aligned.}.
The sequences could be taken at different environments, using different objects, or they may be taken on the same task but using cameras with different viewpoints.
We select one of the images as the anchor, while the image on the other sequence with the same phase index is selected as the positive.
Any other image with a different phase index is a negative candidate.
Although the network is trained with the triplet loss \eqref{eq:tripletloss}, the main feature of using sequential data is that it provides structure to enable self-supervision.
The self-supervision relates to the fact that positives and negatives are given by the indexes of the frames, and thus can be extracted automatically without the need of labels.

Note that in our particular goal of task progress estimation, if the network learns to distinguish the anchor from the adjacent frames, we should expect it to be even more accurate for frames farther apart as the dissimilarity increases as time progresses.
We did not investigate this property in this paper, but it suggests that training of the network can become more efficient by focusing only on the adjacent frames as negatives without affecting accuracy.
Currently, we naively selected the negatives by random sampling frames with different phase indexes from the anchor.

\subsection{Computing a Feedback Policy from a Sequence of Images} \label{sec:policy}
\begin{figure}
	\centering
	\vspace{0.283425cm}
	\includegraphics[width=1.0\linewidth]{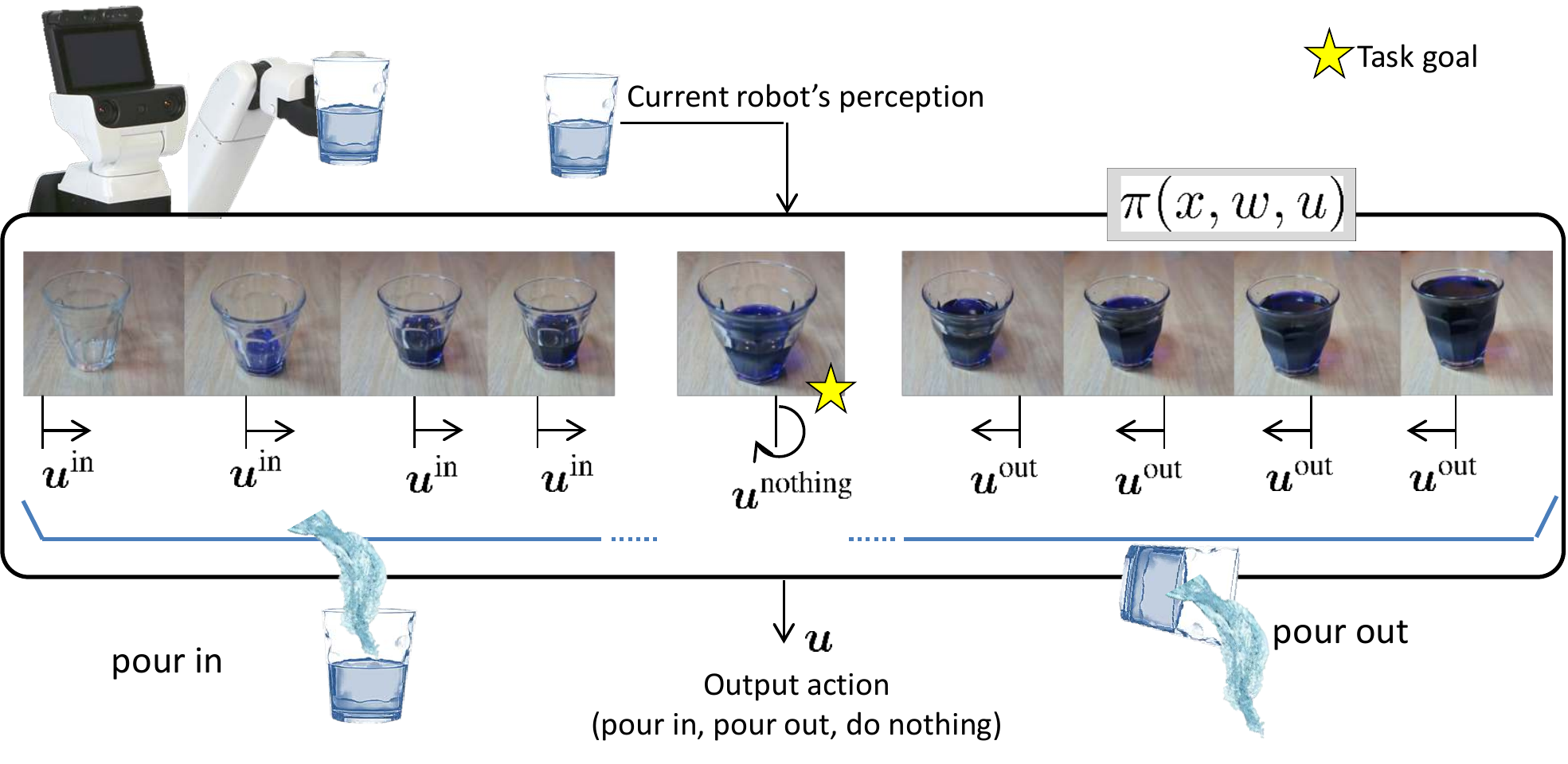}
	\caption{
		The policy consists of associating robot's actions to a sequence of images.
		Given a desired goal image, the direction to which move along the sequence (left, right, none) is found by finding the nearest-neighbor image w.r.t the robot's view of the scene.  
		The output action can be used to trigger a motion planner or as a feedback error signal.
		The photos in the sequence are illustrative and not used for real experiments.}
	\vspace{-0.00cm}
	\label{fig:computing_policy}
\end{figure}

The goal of our policy is to iteratively steer the robot actions such that the desired goal image is achieved in the long-horizon.
We propose the policy as a sequence of the task progress $\pi = \{(x_0, w_0, u_0), ..., (x_N, w_N, u_N )\}$, where each image $x_n$ is associated with an action $u_n$ and embedding $w_n = f(x_n)$.
At execution time, the robot queries its vision sensor for an image $x^*$, compute its embedding $w^*$, and find the closest embedding in the sequence $\pi$ via nearest-neighbor to retrieve the associated action $u^*$.
Since the policy is constrained on a sequence whose frames cannot be skipped (see Figure \ref{fig:computing_policy}) there are only three possible actions for each image: go to the frame on the right, on the left, or do nothing.

Consider a sequence of pouring as shown in Figure \ref{fig:computing_policy}.
Given one of the images in the sequence $\pi$ as the goal state (the starred image), we attribute the images at the right with the action ``pour in'', and the images at the left with ``pour out'' (or ``spread one object'' in a cleaning task).
The action from the current goal image is ``do nothing''.
While during training, an arbitrary number of simultaneous sequences are used (two in the example in Figure \ref{fig:triplet_loss}), at execution time only a single image from the robot's perspective is fed to the network.

The simplicity of the policy is due to two assumptions.
First, the difficult part, which is to disentangle the visual complexity of a scene into a phase index is assumed to be solved by the network.
Second, we assume the robot has means to manipulate the environment such that the next desired state configuration can be achieved, or in other words, that the action induced by the estimated progress can be grounded by appropriate robot motor commands.

The first assumption regards the existence of a properly trained TCN.
The second assumption relies on the existence of a motion planner or a task-and-motion planner (TAMP)\cite{kaelblingHierarchicalTaskMotion2011, dantamIncrementalTaskMotion2016}.
As motivated earlier in the introduction, our goal in training the network is to understand the task state, not to understand the control actions to achieve such states.

\section{Experiments with a Mobile Base Manipulator}

\begin{figure}
	\centering
	\vspace{0.283425cm}
	\includegraphics[width=1.0\linewidth]{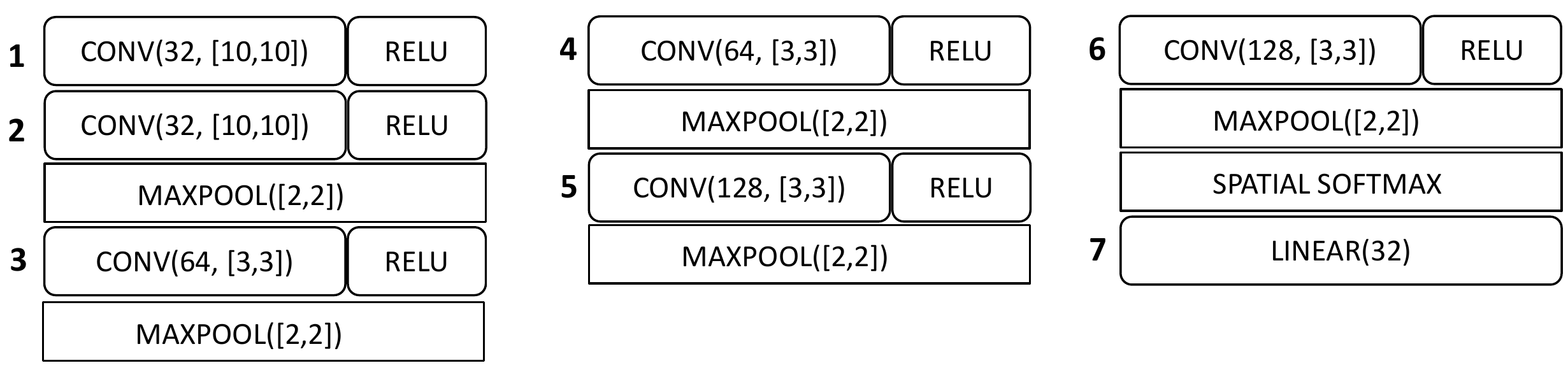}
		\vspace{-0.750cm}
	\caption{
		The network configuration used in both experiments.}
	\label{fig:network}
\end{figure}

Two simulated experiments of different control nature were conducted. 
Both experiments used the same network structure shown in Figure \ref{fig:network} and the same separation margin $\alpha$ of 0.2  but were trained on different data.
In the first experiment, the robot had to either clean up or spread objects inside a specified area using a motion planner.
In the second experiment, the robot had to either fill or empty a simulated cup to a desired level using a bang-bang controller.

We used RGB images of size 300x300 pixels as input to the network.
The network's visual embedder consisted of six padded 2D-convolution layers with (32, 32, 64, 64, 128, 128) channels and 10x10 kernel size on the first layer and 3x3 on the rest. 
Except for the first and last layers, each convolution layer was followed by a 2x2 spatial max-pooling. 
Each layer runs through a ReLu activation function. 
The output of the visual feature extractor was fed to a spatial softmax-function \cite{levineEndtoendTrainingDeep2016} that converts pixel-wise features to feature points. 
Finally, the feature points were fed to a 32-wide MLP and the output embedding vector is L2-normalized.
No batch normalization was used as in the original TCN \cite{sermanet2018time} as we did not notice significant improvements.

While Figure \ref{fig:triplet_loss}(b) illustrated a training with only two sequences, in practice the number of sequences depends on the number of available cameras.
In real-world cases, one sequence most likely is generated by the first-person view of the robot, while another could be obtained from a second camera mounted on the robot's wrist, for example.
In simulation, we set four cameras for training the TCN where their locations are slightly perturbed after each picture is taken except the fourth camera, which is from the robot's simulated sensor.
Thus, one run of data collection generates four sequences.

\begin{figure}
	\centering
	\vspace{0.283425cm}
	\includegraphics[width=0.9\linewidth]{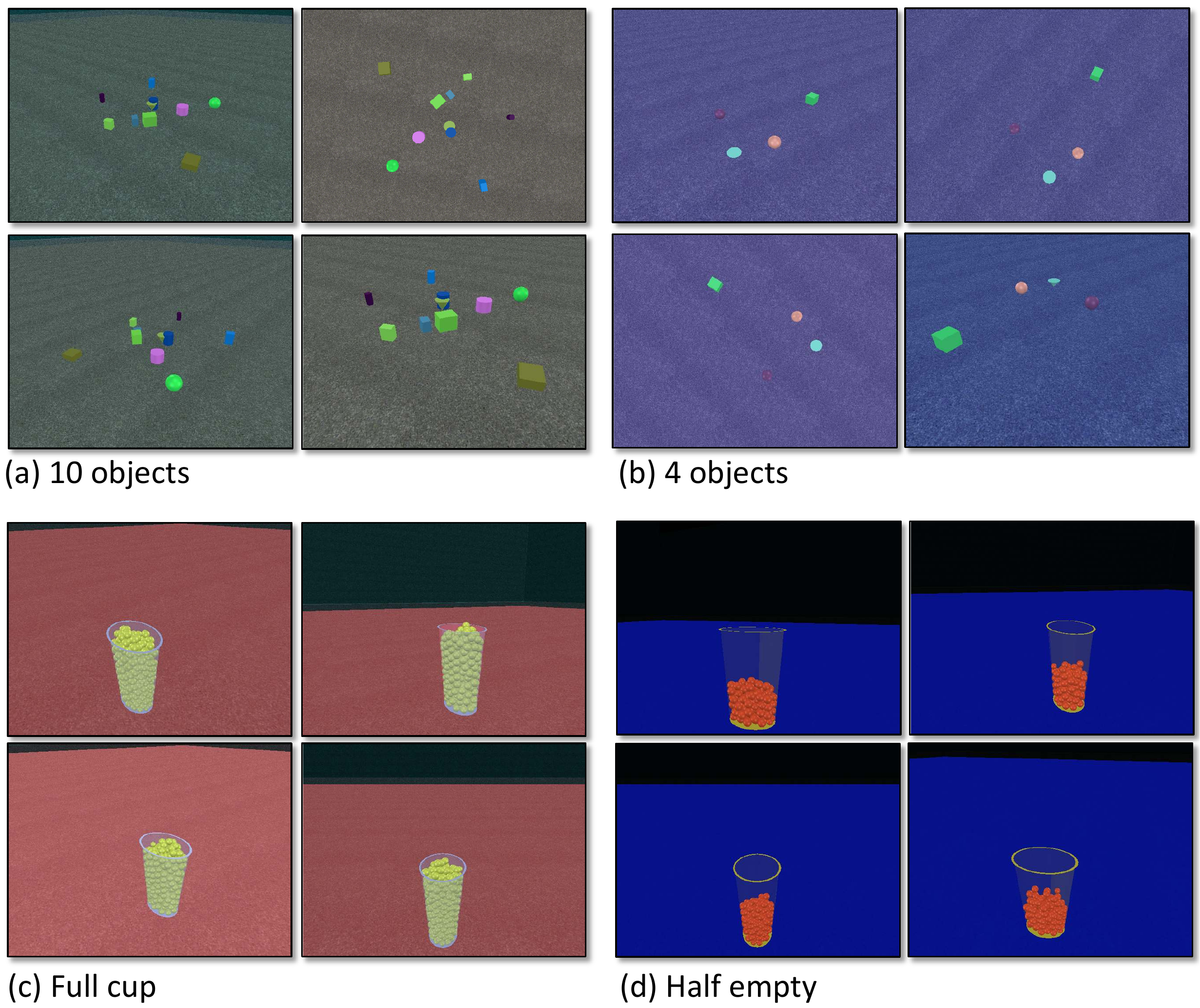}
	\caption{Examples of training data. Each of the four views are capture at the same time-step. The fourth view is the robot's perspective.
		In (a) a cleaning scene with ten objects and in (b) and with four objects.
		(c) Four viewpoints of a full cup (345 particles) and (d) a partially filled cup.
		The pairs (a,b) and (c,d) differ due to the randomization between runs.}
	\label{fig:exampleSequencesTraining}
\end{figure}

\begin{figure}
	\centering
	\vspace{0.283425cm}
	\includegraphics[width=0.9\linewidth]{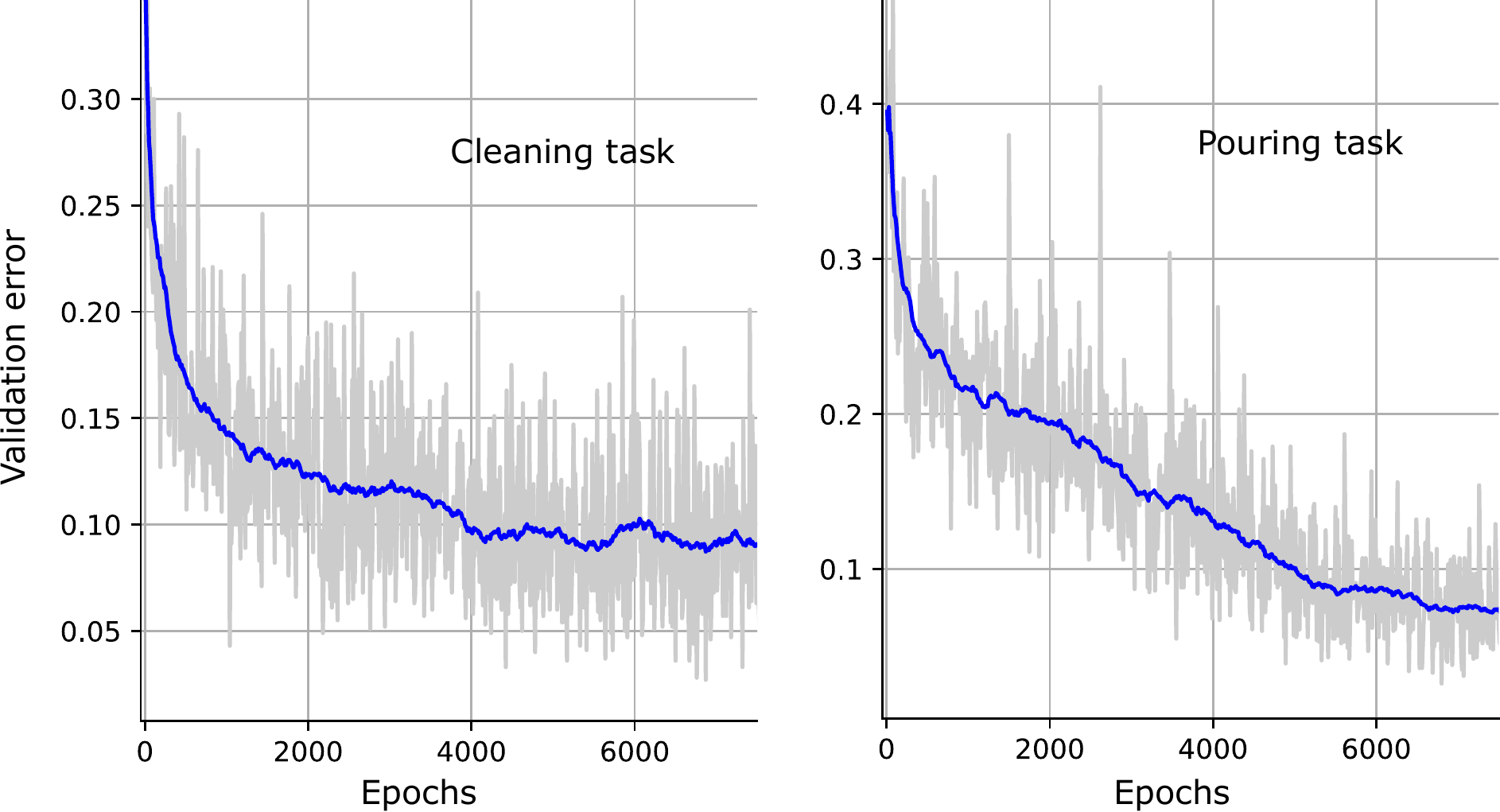}
	\vspace{-0.3cm}
	\caption{
		The decrease in validation error as a function of the number of epochs.
		The total training time was roughly 3 hours in both cases.
	}
	\label{fig:learning_curve}
\end{figure}

For each experiment, a total of 200 randomized runs with variations in color, viewpoint, shape, and size of objects, resulting in 800 sequences were collected for training.
Each sequence was discretized in 16 steps.
Examples of randomized training images are shown in Figure \ref{fig:exampleSequencesTraining} at particular steps of the cleaning (a,b) and pouring (c,d) tasks.
Figure \ref{fig:learning_curve} shows the learning curves on the validation set as a function of the training epochs.
The training time was around three hours in each case on a modest desktop with a single GPU.

\subsection{Cleaning or Spreading Objects on the Floor}

In this experiment we set up a long-horizon task where the robot must either clean a specified floor area $\mathcal{C}$ by moving objects from within this area and placing them in a storage area; or conversely, by spreading objects inside $\mathcal{C}$ by picking them from the storage area.
We used the Toyota Human Support Robot (HSR) 
as the service robot platform.
The robot has an arm with five degrees of freedom and a gripper as an end-effector.
The holonomic base allows the robot to move in XY directions and change its orientation.
Details on the robot can be found in \cite{yamamoto2019development}.

\begin{figure*}
	\centering
	\includegraphics[width=0.9\linewidth]{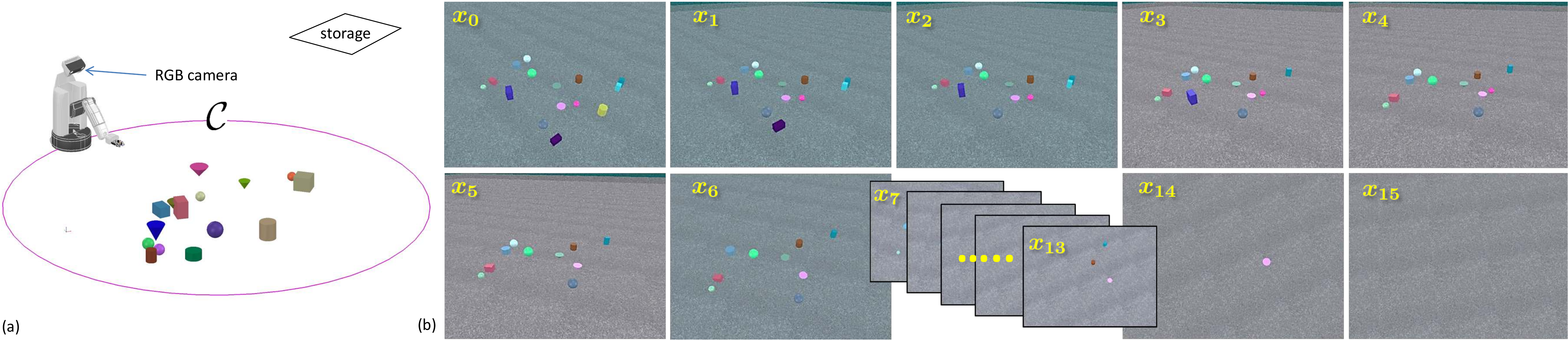}
	\vspace{-0.25cm}    
	\caption{(a) Initial setup of a scene for a long-horizon cleaning task (rendering of the floor is omitted). 
		(b) Sequence of images $\{x_0,...x_{15}\}$  showing one object removed per frame advance.
		This sequence is used to generate the policy $\pi$ from which the robot iteratively retrieves actions until all objects are moved into the storage area.}
	\label{fig:exp_clean_floor_setup}
\end{figure*}

Figure \ref{fig:exp_clean_floor_setup}(a) shows the scene setup at the start of a cleaning task\footnote{the texture of the floor was suppressed in the figure to facilitate the visualization of the scene, but is present on the data fed to the network.}.
Figure \ref{fig:exp_clean_floor_setup}(b) shows the sequence of 16 images $\{x_0, ..., x_{15}\}$ used to compute the policy $\pi$ where 15 objects of random shapes and sizes are initially provided and removed one-by-one at each frame advance.
Note that due to the random spreading of objects, scenes where objects occluded each other were also included in the training and test data.
If the task is to clean the floor, we specify the goal of the task by showing the robot the image $x_{15}$. 
If the task is to spread blocks inside the area delimited by $\mathcal{C}$, we specify the goal of the task as the image $x_{0}$ and initialize the scene with 15 blocks in the storage area.
Any other intermediate stage can also be used as a goal.
After a goal is specified, the actions in the policy $\pi$ are computed as described in Sec. \ref{sec:policy}.

\begin{figure*}
	\centering
	\includegraphics[width=0.9\linewidth]{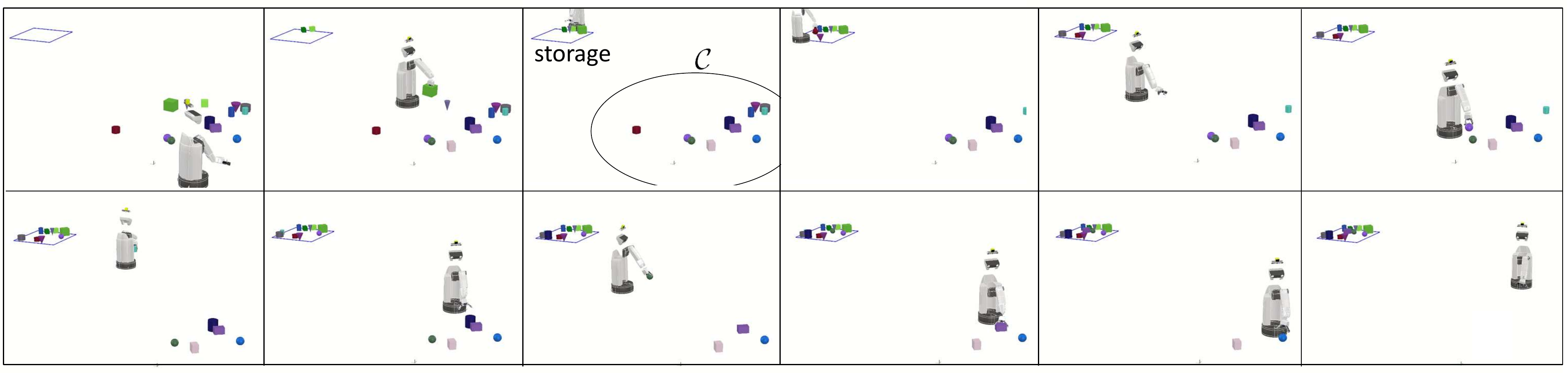}
	\vspace{-0.15cm}
	\caption{Sequence of snapshots taken at varying steps where the robot executes an entire clean-up task. 
		A motion planner is used to ground the high-level commands ``remove object'', ``add object'', ``do nothing'' obtained from the policy $\pi(x,w,u)$.
		The borders of the area $\mathcal{C}$ is only shown in the third snapshot and are not visible to the robot's vision sensor. The rendering of the floor was omitted for clarity but it is visible to the robot.
		An extended video of experiments can be watched in the link \url{https://youtu.be/t5WkdECAD1g}.	
	}
	\label{fig:sequence_example}    
\end{figure*}

Figure \ref{fig:sequence_example} shows snapshots of an entire cleaning task where the motion planner is used to ground the high-level actions $u$.
In the first frame, the robot retrieves images from the scene $x^*$ using its simulated RGB camera from the perimeter of $\mathcal{C}$. 
The coordinates to which the robot moves on the border $(x_c,y_c) \in \mathcal{C}$ are randomly picked such that the robot has no pre-defined position from which to observe the scene.
From the image, the embedding $w^* = f(x^*)$ is computed, and the appropriate actions are retrieved by finding the nearest-neighbor in $\pi$.

If the retrieved action $u^*$ is to ``remove object'' the robot picks the closest object to itself (second snapshot of Figure \ref{fig:sequence_example}) and drops it inside the storage area (third snapshot).
If the action is ``spread one object'' the robot picks the object from the storage area and places it randomly within the boundaries delimited by $\mathcal{C}$.
Once either of these actions is taken, the robot goes back to the perimeter of $\mathcal{C}$ (fifth snapshot) to obtain a new image and retrieve the next action. 
The robot repeats the process until the nearest-neighbor in the embedding space reaches the specified goal image.
From that point on, any extra step will generate a ``do nothing'' action (last snapshot). 
A video on the experiment is available as accompanying material\footnote{A high-resolution version of the video is accessible at \url{https://youtu.be/t5WkdECAD1g}.}.

\begin{figure}
	\centering
	\includegraphics[width=0.8\linewidth]{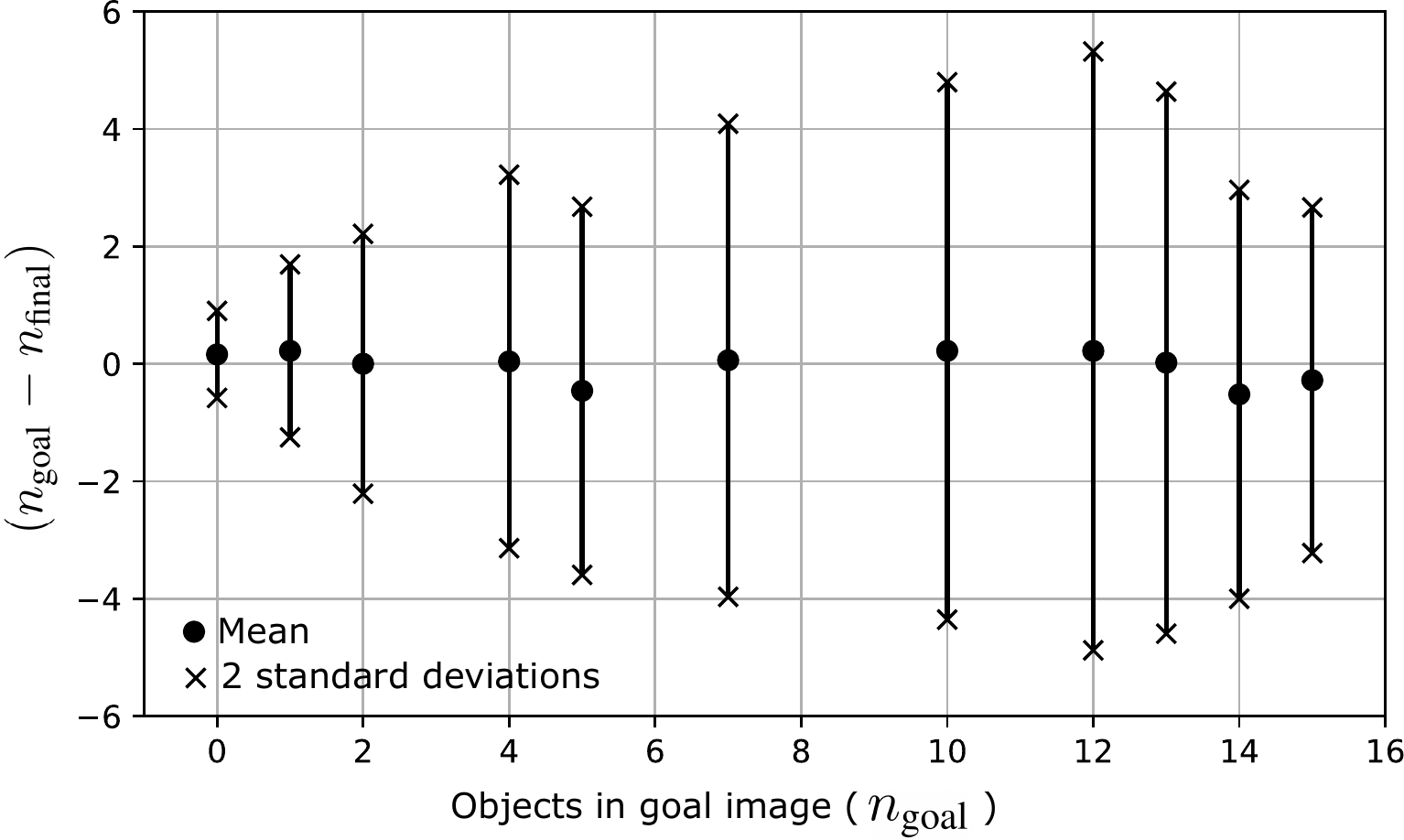}
	\vspace{-0.25cm}
	\caption{Error in the number of objects in the scene after the task was considered finished by the robot. The precision is high when completely cleaning the floor but tends to worsen as clutter increases.
	}
	\label{fig:clean_accuracy}
\end{figure}

Simulations allow for quantification against a ground-truth.
The goal state is achieved if the final number of objects in the scene 
$n_{\text{final}}$
is the same as the number of objects in the goal image $n_{\text{goal}}$.
An error in task execution means that the robot reaches the ``do nothing'' action despite the number of objects in the final scene and goal image being different.
Figure \ref{fig:clean_accuracy} shows the final error where each error bar summarizes 50 tasks under randomized initial conditions. 
The X-axis represents the number of objects in the goal image $n_{\text{goal}}$.
For example, when $n_{\text{goal}}$ is zero, the goal is to completely clean the floor.
Since the maximum number of objects inside  $\mathcal{C}$ was 15, there were 14 possible initializations regarding the number of objects at the beginning of the task.

Figure \ref{fig:clean_accuracy} shows that the error is reasonably constant and nearly zero throughout the range of goal objects.
On the other hand, the magnitude of the error bars changes according to $n_{\text{goal}}$ indicating that the precision is affected by the clutter of the goal images.
The robot achieves the lowest variance when the goal is to completely wipe the floor. 
This result is somewhat intuitive as the scene without objects is the most visually discernible  among its adjacent frames.
On the other hand, the precision is worse when the clutter is large ($10 < n_{\text{goal}} < 13$) which also reflects our difficulty in visually differentiating a scene with 10 objects from a scene with 11 objects.
The precision improves slightly when nearing $n_{\text{goal}}=$15 as the storage area is depleted.

For us, it is obvious that the underlying operation in the proposed task is to count the number of objects.
It may seem surprising that a network can achieve similar counting capability from images although it was not explicitly trained to count and neither has a concept of number.
However, recall that task progress, although not visible to the human eye, is the underlying mechanism for the network training.
The network is sensitive to phases, which in turn can be used by the robot to identify how many steps, or objects, have been manipulated given the policy as a sequence of images.
As such, when discretizing a sequence in 16 steps for training, we must assume the same steps across all sequences represent the same phase.

\begin{figure}
	\centering
	\includegraphics[width=0.7\linewidth]{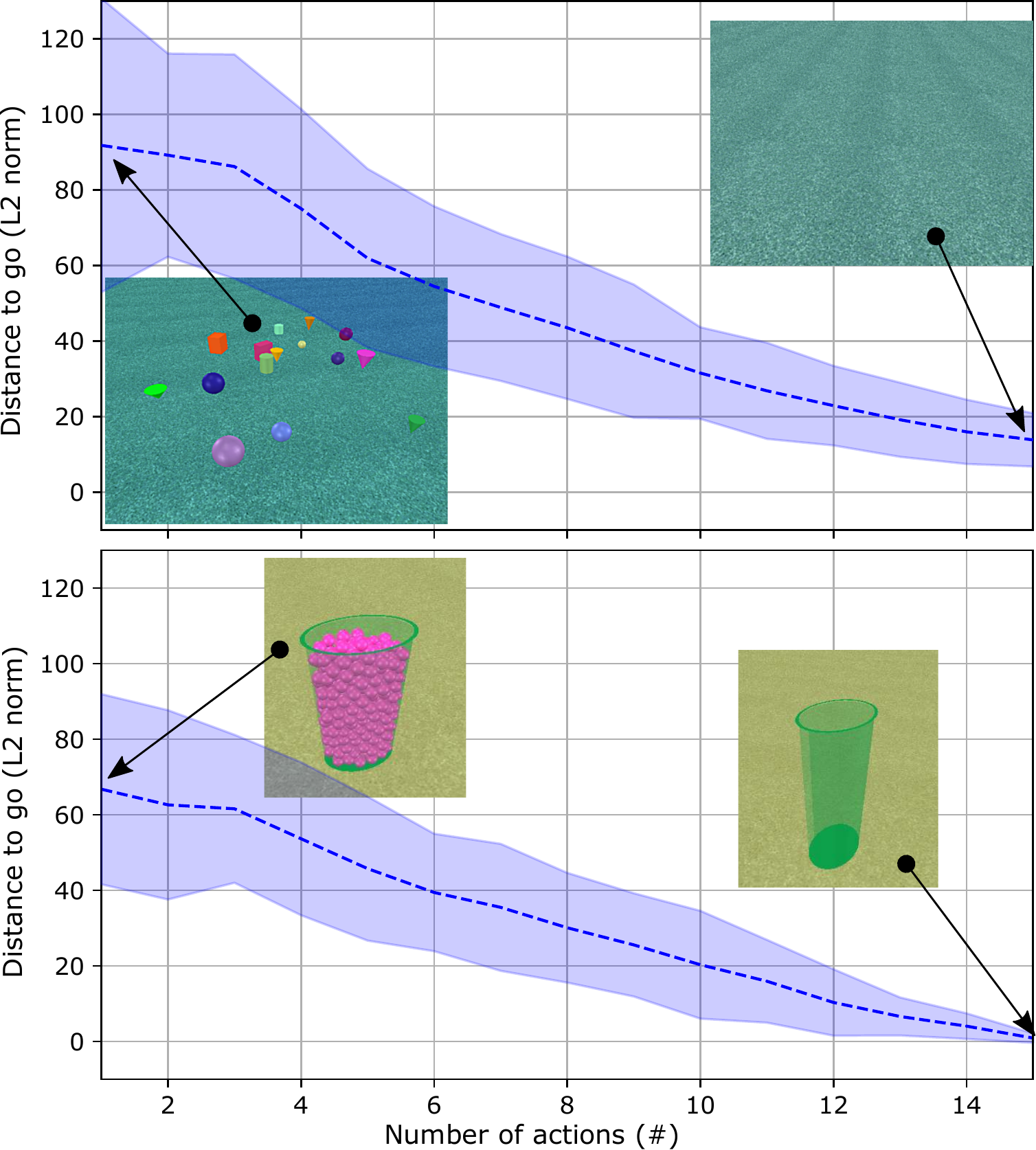}
	\vspace{-0.25cm}
	\caption{History of distance-to-go as mean and two standard deviations on the cleaning task (top) and pouring (bottom).
		Both tasks were discretized in 15 steps. Meaning, in the cleaning case 1 block is removed per frame advance, and in the pouring task 23 particles are decreased per frame.
		The distance values are respective to the last frame of the sequence (clean floor and empty cup).
	}
	\label{fig:block_traj_cost_clean_and_spread}
\end{figure}

Figure \ref{fig:block_traj_cost_clean_and_spread} shows distributions containing 50 ``distance-to-go'' trajectories, that is, trajectories representing the progress of the task in relation to a final desired state.
All values in the Y-axis are distances relative to the goal image of a floor without blocks (top plot) and an empty cup (bottom plot).
In the cleaning case, the robot starts with all the 15 blocks in front of it, and the goal is to completely clean the floor following the policy $\pi$.
Note that, in general, the L2 distance decreases as the goal is approached.
At first glance, the fact that the distance decreases in an almost monotonic fashion is somewhat odd as the negatives in the triplet-loss 
do not contain explicit information regarding distance towards the anchor.
This phase-driven dissimilarity indicates that training with a triplet-loss under time-varying phenomena can be, in fact, easier than training static cases where data-mining to search for hard cases is an actual issue \cite{schroffFacenetUnifiedEmbedding2015}.

\subsection{Pouring in a Simulated Cup by Directly Looking at a Screen}
\begin{figure}
	\centering
	\includegraphics[width=0.7\linewidth]{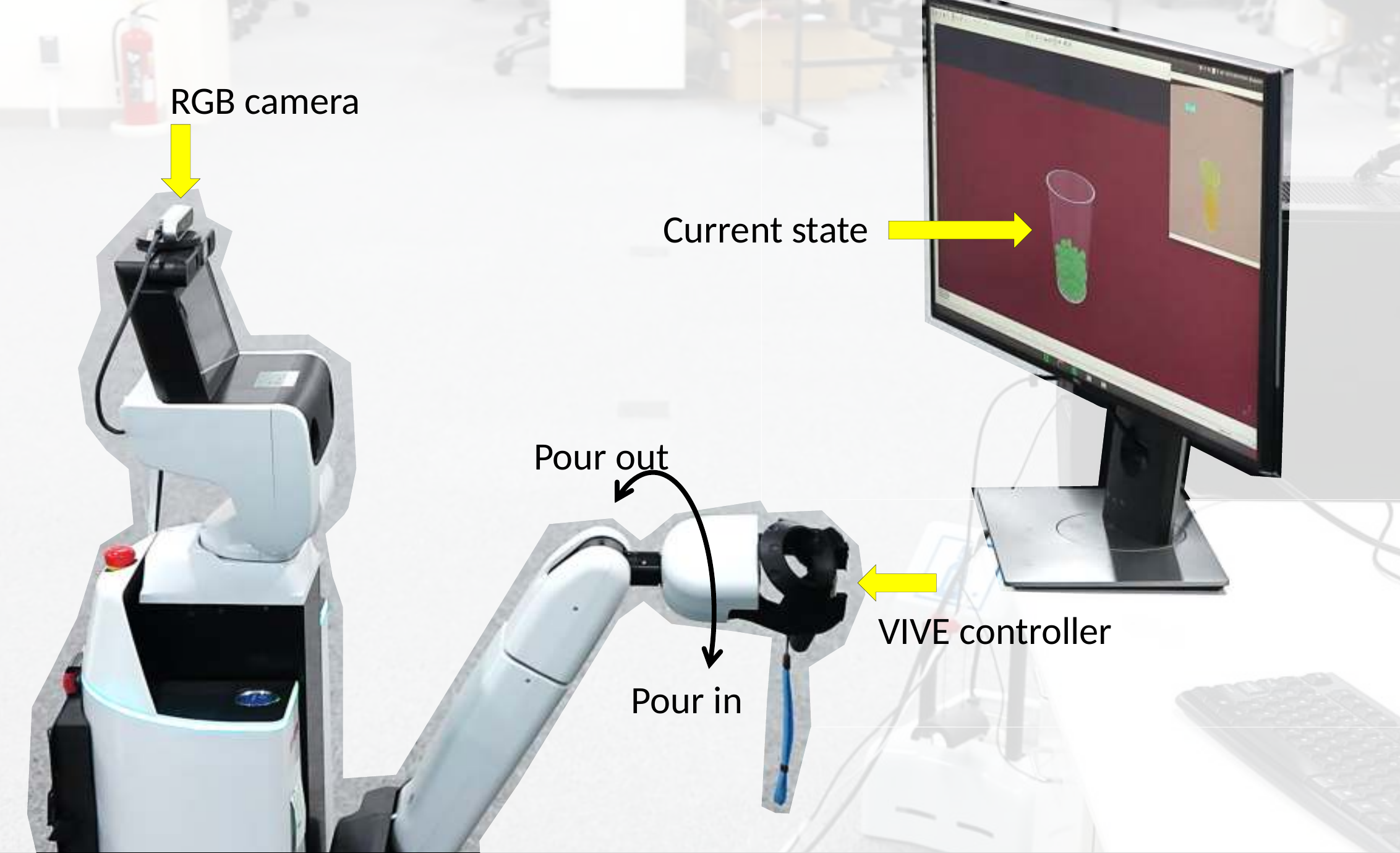}
	\vspace{-0.25cm}
	\caption{Cup setup experiment where the robot interacts with a simulator via an RGB camera at the top of its head and a VIVE controller.
		The robot controls the pouring amount by rotating the controller while observing the state of the cup by looking directly at the monitor.}
	\label{fig:exp_cup_setup}
\end{figure}

Here, we use the visual policy $\pi$ as a feedback error signal and directly feed it to a bang-bang control law (pour-in/pour out).
Figure \ref{fig:exp_cup_setup} shows the hybrid real-virtual experimental setup where particles are used as a proxy of liquid for ground truth purposes.
A VIVE controller is used to send commands to a simulator, which detects the inclination of the controller.
When the controller is turned more than 35$^0$, 15 particles fall inside the cup (4$\%$ of the total volume of a full cup).
When turned less than $-$35$^0$, 15 particles are removed from the cup.
The network is trained on images directly collected from the simulator (as shown in Figure \ref{fig:exampleSequencesTraining}) but at run-time, the image is obtained from the robot's RGB camera pointed at the monitor.
As such, the images are cropped to eliminate the physical borders of the monitor\footnote{This arrangement brings us close to a full real experiment in which we are currently working to replace the VIVE controller with a real cup.}

Images in the simulator are randomized in regards to the color and transparency of the particles, the cup,  the background, and variations in camera viewpoint. 
The geometry of the cup is modified by scaling one of the principal axes of the rim (from circular to ellipsoidal) and the height.
Both scales vary from 1 to 1.45.
Note that, different from the previous experiment, here the viewpoint of the robot is fixed as the robot constantly looks at the cup in its hand.
Thus, to increase geometry variation on the robot's view, before each trial, the cup is randomly rotated along its vertical axis such that the principal axis of the ellipsoidal cup is unlikely to repeat directions.

\begin{figure}
	\centering
	\includegraphics[width=0.8\linewidth]{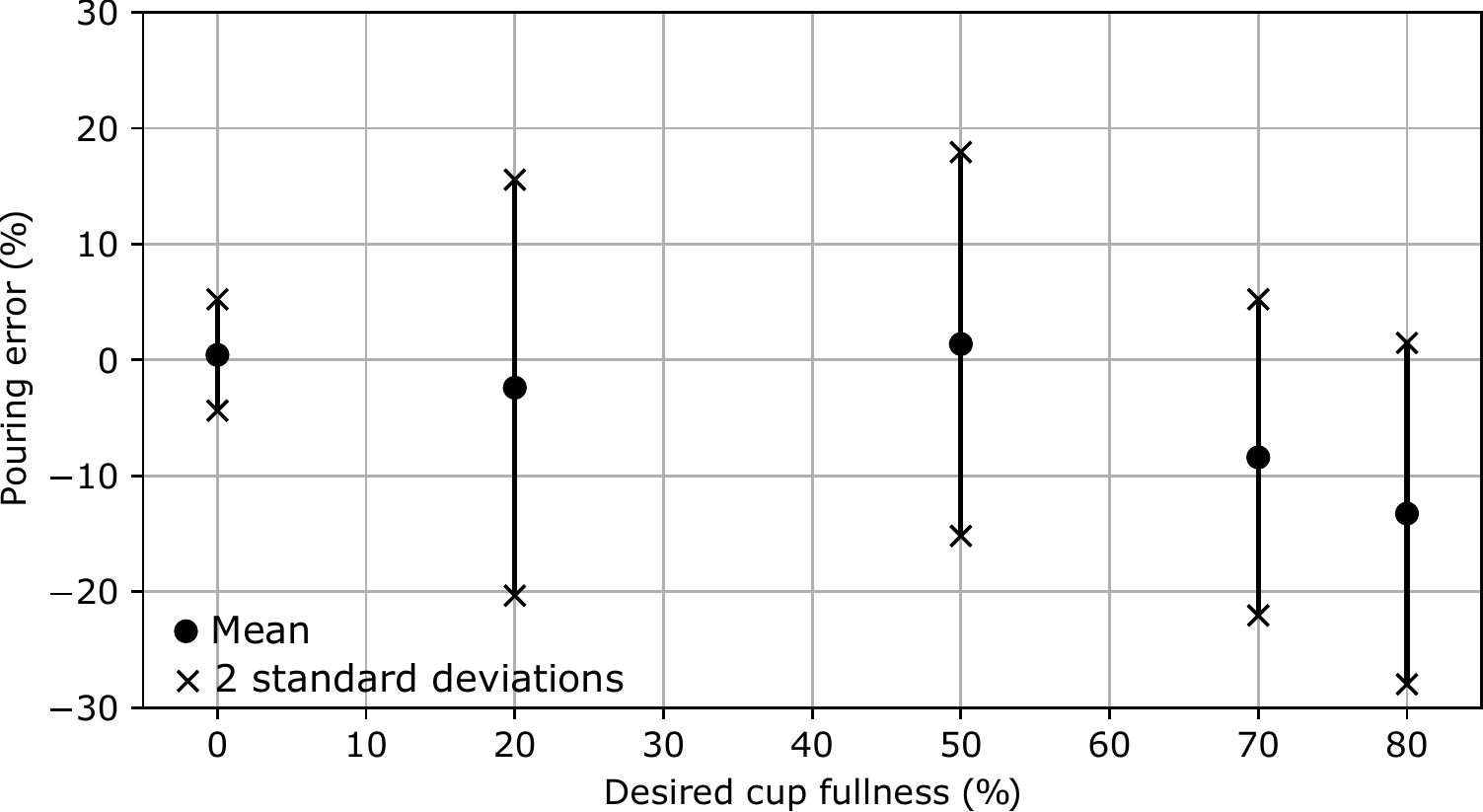}
	\vspace{-0.3cm}
	\caption{
		Cup pouring accuracy according to different levels of desired particles. The variance is reasonably constant but there is a trend to under-pour as the desired volume increases. We conjecture this may be due to the unfavorable angle from which the robot observes the cup.
	}
	\label{fig:cup_fill_accuracy}
\end{figure}

Figure \ref{fig:cup_fill_accuracy} quantifies the pouring error as a function of the amount of desired ``liquid'' inside the cup.
On the X-axis, the desired fullness of 0$\%$ is represented by an image of an empty cup.
The desired fullness of 80$\%$ is given by an image of a cup with 275 particles.
The Y-axis shows the mean and $\pm$ two standard deviations from 20 trials per case.
Apart from the case where the goal is to empty the cup, the variance is roughly similar.
It is observed a trend, however, where the robot tends to fill less of the cup as the amount of desired liquid increases.
It is not clear why this trend occurs, but it could be related to the fact that the fixed angle from which the robot observed the cup does not favor such assessment (that is, the best angle to assess the liquid level would be to look from the side of the cup, and worst case to look from the top).
In the case of the cleaning task, this effect was not noticeable as the robot's view of the scene included all objects in a single image.

\begin{figure*}
	\centering
	\includegraphics[width=0.8\linewidth]{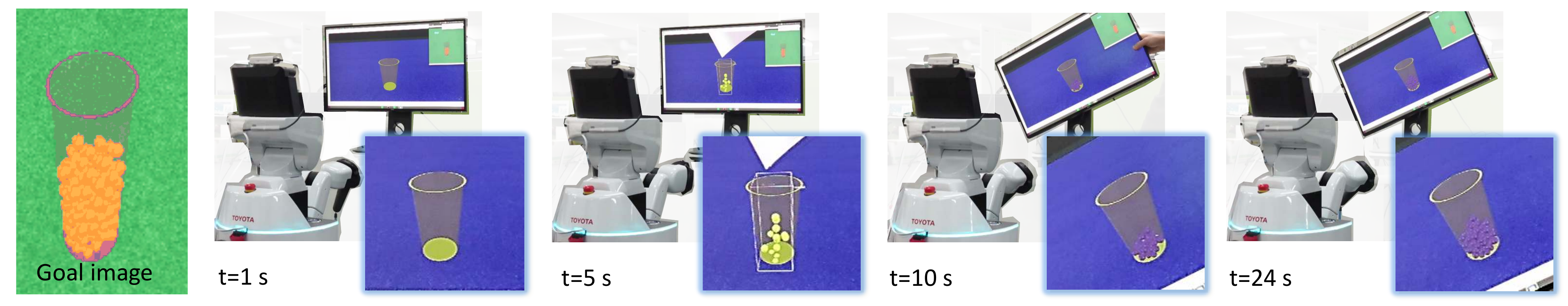}
	\vspace{-0.25cm}
	\caption{
		One instance of the pouring task.
		The image at the left is the goal image. The sequence of snapshots shows the robot adding particles into the cup. 
		The monitor is physically rotated during the task.
		The robot achieves a similar volume to the goal despite the visual differences with the goal image and the tilt of the monitor (the tilt was never seen during training).
	}
	\label{fig:sequence_example_cup}
\end{figure*}

Figure \ref{fig:sequence_example_cup} shows snapshots of one instance of the experiment where the desired volume to be attained is given by the image at the left.
The robot rotates the controller, which increases the volume by 15 particles in its cup and returns the controller to its original $0^0$ rotation.
The robot then observes the monitor and searches for the next action in the policy.
During the experiment (see accompanying video) the monitor is physically tilted and the robot could still reasonably accomplish the task (20$\%$ pouring error) although the embeddings were not trained on tilted images and images observed from a real camera.

\section{Directions for Future Work}

We trained the network naively by random sampling negatives in the triplet loss.
We believe there are more efficient ways to train temporal triplet loss networks by focusing the learning only on the adjacent frames as negatives.
As discussed when presenting Figure \ref{fig:triplet_loss}(b), if the network learns to separate adjacent frames it should automatically learn to separate the remaining frames as well.

The literature for temporal learning and representation learning with multiple-views is extremely rich.
Multi-modal learning \cite{aytarPlayingHardExploration2018}, 
learning from mutual information across views  \cite{bachmanLearningRepresentationsMaximizing2019}, using prediction for learning features in self-supervision  \cite{oordRepresentationLearningContrastive2019}, etc. are just a few of the approaches that have been  proposed outside the robotics scope.
For this paper, TCNs were a natural fit as a method designed specifically for robotics, but we leave for future work a benchmarking among pure vision methods for temporal learning or multiple-view learning, which could potentially be used as an alternative to TCNs in the method here proposed.

We are currently working on a real-world implementation integrating SLAM, planning and collision-avoidance, object instance segmentation and pose estimation for grasping. 
Although more complex than the simulated system, we expect to run the same network and time-contrastive network training procedure previously presented.

\section{Conclusion}

In this paper, we re-visited Time-Contrastive Networks and introduced it as a method to specialize in the degree of accomplishment of a task (that is, the progress) rather than on robot motor skills.
With our formulation, the robot can compute distances towards task accomplishment regardless of the appearance of images.
To ground the estimated progress with physical robot actions, we introduced a simple policy based on a sequence of images representing gradual task accomplishment, where each frame is associated with an action.
These actions were evaluated both as a task specification for long-horizon task planning and as a feedback error signal for low-level control.

\section{Acknowledgment}

The authors would like to acknowledge Koichi Ikeda, Kunihiro Iwamoto and Takashi Yamamoto from Toyota Motor Corporation for their support of the HSR platforms.

\bibliographystyle{IEEEtran}
\bibliography{./TCN_related_work2}

\begin{thebibliography}{10}
\providecommand{\url}[1]{#1}
\csname url@samestyle\endcsname
\providecommand{\newblock}{\relax}
\providecommand{\bibinfo}[2]{#2}
\providecommand{\BIBentrySTDinterwordspacing}{\spaceskip=0pt\relax}
\providecommand{\BIBentryALTinterwordstretchfactor}{4}
\providecommand{\BIBentryALTinterwordspacing}{\spaceskip=\fontdimen2\font plus
\BIBentryALTinterwordstretchfactor\fontdimen3\font minus
  \fontdimen4\font\relax}
\providecommand{\BIBforeignlanguage}[2]{{%
\expandafter\ifx\csname l@#1\endcsname\relax
\typeout{** WARNING: IEEEtran.bst: No hyphenation pattern has been}%
\typeout{** loaded for the language `#1'. Using the pattern for}%
\typeout{** the default language instead.}%
\else
\language=\csname l@#1\endcsname
\fi
#2}}
\providecommand{\BIBdecl}{\relax}
\BIBdecl

\bibitem{sermanet2018time}
P.~Sermanet, C.~Lynch, Y.~Chebotar, J.~Hsu, E.~Jang, S.~Schaal, S.~Levine, and
  G.~Brain, ``Time-contrastive networks: {Self}-supervised learning from
  video,'' in \emph{2018 {IEEE} international conference on robotics and
  automation ({ICRA})}, 2018, pp. 1134--1141, tex.organization: IEEE.

\bibitem{kuniyoshiLearningWatchingExtracting1994}
Y.~Kuniyoshi, M.~Inaba, and H.~Inoue, ``Learning by watching: {Extracting}
  reusable task knowledge from visual observation of human performance,''
  \emph{Robotics and Automation, IEEE Transactions on}, vol.~10, no.~6, pp.
  799--822, 1994.

\bibitem{bentivegnaLearningTasksObservation2004}
D.~C. Bentivegna, C.~G. Atkeson, and G.~Cheng, ``Learning tasks from
  observation and practice,'' \emph{Robotics and Autonomous Systems}, vol.~47,
  no.~2, pp. 163--169, 2004.

\bibitem{finnDeepSpatialAutoencoders2016}
C.~Finn, X.~Y. Tan, Y.~Duan, T.~Darrell, S.~Levine, and P.~Abbeel, ``Deep
  spatial autoencoders for visuomotor learning,'' in \emph{2016 {IEEE}
  {International} {Conference} on {Robotics} and {Automation} ({ICRA})}.\hskip
  1em plus 0.5em minus 0.4em\relax IEEE, 2016, pp. 512--519.

\bibitem{sermanetUnsupervisedPerceptualRewards2017}
P.~Sermanet, K.~Xu, and S.~Levine, ``Unsupervised perceptual rewards for
  imitation learning,'' in \emph{Proceedings of robotics: {Science} and
  systems}, Cambridge, Massachusetts, Jul. 2017.

\bibitem{nairVisualReinforcementLearning2018}
A.~V. Nair, V.~Pong, M.~Dalal, S.~Bahl, S.~Lin, and S.~Levine, ``Visual
  reinforcement learning with imagined goals,'' in \emph{Advances in {Neural}
  {Information} {Processing} {Systems}}, 2018, pp. 9191--9200.

\bibitem{Pinto-RSS-18}
L.~Pinto, M.~Andrychowicz, P.~Welinder, W.~Zaremba, and P.~Abbeel, ``Asymmetric
  actor critic for image-based robot learning,'' in \emph{Proceedings of
  robotics: {Science} and systems}, Pittsburgh, Pennsylvania, Jun. 2018.

\bibitem{jang18a_grasp2vec}
E.~Jang, C.~Devin, V.~Vanhoucke, and S.~Levine, ``{Grasp2Vec}: {Learning}
  object representations from self-supervised grasping,'' in \emph{Proceedings
  of the 2nd conference on robot learning}, ser. Proceedings of machine
  learning research, vol.~87.\hskip 1em plus 0.5em minus 0.4em\relax PMLR, Oct.
  2018, pp. 99--112.

\bibitem{schroffFacenetUnifiedEmbedding2015}
F.~Schroff, D.~Kalenichenko, and J.~Philbin, ``Facenet: {A} unified embedding
  for face recognition and clustering,'' in \emph{Proceedings of the {IEEE}
  conference on computer vision and pattern recognition}, 2015, pp. 815--823.

\bibitem{hermansDefenseTripletLoss2017a}
A.~Hermans, L.~Beyer, and B.~Leibe, ``In defense of the triplet loss for person
  re-identification,'' \emph{arXiv preprint arXiv:1703.07737}, 2017.

\bibitem{schmidtSelfsupervisedVisualDescriptor2016}
T.~Schmidt, R.~Newcombe, and D.~Fox, ``Self-supervised visual descriptor
  learning for dense correspondence,'' \emph{IEEE Robotics and Automation
  Letters}, vol.~2, no.~2, pp. 420--427, 2016.

\bibitem{zeng3dmatchLearningLocal2017}
A.~Zeng, S.~Song, M.~Nießner, M.~Fisher, J.~Xiao, and T.~Funkhouser,
  ``3dmatch: {Learning} local geometric descriptors from rgb-d
  reconstructions,'' in \emph{Proceedings of the {IEEE} {Conference} on
  {Computer} {Vision} and {Pattern} {Recognition}}, 2017, pp. 1802--1811.

\bibitem{florenceDenseObjectNets2018}
P.~R. Florence, L.~Manuelli, and R.~Tedrake, ``Dense object nets: {Learning}
  dense visual object descriptors by and for robotic manipulation,''
  \emph{arXiv preprint arXiv:1806.08756}, 2018.

\bibitem{florenceSelfSupervisedCorrespondenceVisuomotor2019}
P.~Florence, L.~Manuelli, and R.~Tedrake, ``Self-supervised correspondence in
  visuomotor policy learning,'' \emph{IEEE Robotics and Automation Letters},
  2019, tex.publisher: IEEE.

\bibitem{dwibedi2018learning}
D.~Dwibedi, J.~Tompson, C.~Lynch, and P.~Sermanet, ``Learning actionable
  representations from visual observations,'' in \emph{2018 {IEEE}/{RSJ}
  {International} {Conference} on {Intelligent} {Robots} and {Systems}
  ({IROS})}, 2018, pp. 1577--1584, tex.organization: IEEE.

\bibitem{jeongSelfSupervisedSimtoRealAdaptation2019}
R.~Jeong, Y.~Aytar, D.~Khosid, Y.~Zhou, J.~Kay, T.~Lampe, K.~Bousmalis, and
  F.~Nori, ``Self-{Supervised} {Sim}-to-{Real} {Adaptation} for {Visual}
  {Robotic} {Manipulation},'' \emph{arXiv:1910.09470 [cs]}, Oct. 2019.

\bibitem{liuImitationObservationLearning2018}
Y.~Liu, A.~Gupta, P.~Abbeel, and S.~Levine, ``Imitation from observation:
  {Learning} to imitate behaviors from raw video via context translation,'' in
  \emph{2018 {IEEE} {International} {Conference} on {Robotics} and {Automation}
  ({ICRA})}.\hskip 1em plus 0.5em minus 0.4em\relax IEEE, 2018, pp. 1118--1125.

\bibitem{levineEndtoendTrainingDeep2016}
S.~Levine, C.~Finn, T.~Darrell, and P.~Abbeel, ``End-to-end training of deep
  visuomotor policies,'' \emph{Journal of Machine Learning Research}, vol.~17,
  no.~39, pp. 1--40, 2016.

\bibitem{yuOneshotImitationObserving2018}
T.~Yu, C.~Finn, A.~Xie, S.~Dasari, T.~Zhang, P.~Abbeel, and S.~Levine,
  ``One-shot imitation from observing humans via domain-adaptive
  meta-learning,'' \emph{arXiv preprint arXiv:1802.01557}, 2018.

\bibitem{laumondKineoCAMSuccess2006}
J.-P. Laumond, ``\BIBforeignlanguage{en}{Kineo {CAM}: a success story of motion
  planning algorithms},'' \emph{\BIBforeignlanguage{en}{IEEE Robotics \&
  Automation Magazine}}, vol.~13, no.~2, pp. 90--93, Jun. 2006.

\bibitem{krogerOpeningDoorNew2011}
T.~Kröger, ``Opening the door to new sensor-based robot applications—{The}
  {Reflexxes} {Motion} {Libraries},'' in \emph{2011 {IEEE} {International}
  {Conference} on {Robotics} and {Automation}}.\hskip 1em plus 0.5em minus
  0.4em\relax IEEE, 2011, pp. 1--4.

\bibitem{dantamIncrementalTaskMotion2016}
N.~T. Dantam, Z.~K. Kingston, S.~Chaudhuri, and L.~E. Kavraki, ``Incremental
  {Task} and {Motion} {Planning}: {A} {Constraint}-{Based} {Approach},''
  \emph{Robotics: Science and Systems}, 2016.

\bibitem{kaelblingHierarchicalTaskMotion2011}
L.~P. Kaelbling and T.~Lozano-Pérez, ``Hierarchical task and motion planning
  in the now,'' in \emph{Robotics and {Automation} ({ICRA}), 2011 {IEEE}
  {International} {Conference} on}.\hskip 1em plus 0.5em minus 0.4em\relax
  IEEE, 2011, pp. 1470--1477.

\bibitem{yamamoto2019development}
T.~Yamamoto, K.~Terada, A.~Ochiai, F.~Saito, Y.~Asahara, and K.~Murase,
  ``Development of {Human} {Support} {Robot} as the research platform of a
  domestic mobile manipulator,'' \emph{ROBOMECH journal}, vol.~6, no.~1, p.~4,
  2019, tex.publisher: Springer.

\bibitem{aytarPlayingHardExploration2018}
Y.~Aytar, T.~Pfaff, D.~Budden, T.~Paine, Z.~Wang, and N.~de~Freitas, ``Playing
  hard exploration games by watching youtube,'' in \emph{Advances in {Neural}
  {Information} {Processing} {Systems}}, 2018, pp. 2930--2941.

\bibitem{bachmanLearningRepresentationsMaximizing2019}
P.~Bachman, R.~D. Hjelm, and W.~Buchwalter, ``Learning representations by
  maximizing mutual information across views,'' in \emph{Advances in neural
  information processing systems}, 2019, pp. 15\,509--15\,519.

\bibitem{oordRepresentationLearningContrastive2019}
A.~v.~d. Oord, Y.~Li, and O.~Vinyals, ``Representation {Learning} with
  {Contrastive} {Predictive} {Coding},'' \emph{arXiv:1807.03748 [cs, stat]},
  Jan. 2019, arXiv: 1807.03748.

\end{thebibliography}

\end{document}